\pdfoutput=1
\documentclass{article} 
\usepackage{iclr2027_conference,times}

\usepackage{amsmath,amsfonts,bm}

\def\eqref#1{equation~\ref{#1}}

\def\1{\bm{1}}

\DeclareMathAlphabet{\mathsfit}{\encodingdefault}{\sfdefault}{m}{sl}
\SetMathAlphabet{\mathsfit}{bold}{\encodingdefault}{\sfdefault}{bx}{n}

\usepackage{graphicx}
\usepackage{xcolor}
\usepackage{colortbl}
\usepackage{array}
\usepackage{pifont}
\usepackage{caption}
\usepackage{subcaption}
\usepackage{booktabs,multirow}
\usepackage{wrapfig}
\usepackage{tcolorbox}
\tcbuselibrary{breakable, skins}

\usepackage[T1]{fontenc}    
\usepackage{hyperref}       
\usepackage{url}            
\usepackage{amsfonts}       
\usepackage{nicefrac}       
\usepackage{microtype}      
\usepackage{amsmath}

\definecolor{plotblue}{HTML}{1f77b4}
\definecolor{plotorange}{HTML}{ff7f0e}
\definecolor{plotgreen}{HTML}{2ca02c}
\newsavebox{\aucbox}

\title{LLM-Guided Dynamic Action Spaces for Synthesizable Molecular Optimization}

\author{%
  \normalfont
  \makebox[\textwidth][c]{%
    \textbf{Tao Li}\textsuperscript{1},
    \textbf{Kaiyuan Hou}\textsuperscript{2},
    \textbf{Tuan Vinh}\textsuperscript{3},
    \textbf{Fanglei Xue}\textsuperscript{4},
    \textbf{Monika Raj}\textsuperscript{5},
    \textbf{Zhichun Guo}\textsuperscript{6},
    and \textbf{Carl Yang}\textsuperscript{1}%
    \thanks{Corresponding author: \texttt{jyang71@emory.edu}}
  } \\[8pt]
  \makebox[\textwidth][c]{%
    \hspace*{-0.22\textwidth}%
    \parbox{0.82\textwidth}{%
      \raggedright
      \textsuperscript{1}Department of Computer Science, Emory University\\
      \textsuperscript{2}School of Computer Science, Carnegie Mellon University\\
      \textsuperscript{3}MRC Brain Network Dynamics Unit, University of Oxford\\
      \textsuperscript{4}Department of Biochemistry, University of Washington\\
      \textsuperscript{5}Department of Chemistry, Emory University\\
      \textsuperscript{6}Independent Researcher
    }%
  }
}

\iclrfinalcopy 

\begin{document}

\maketitle

\lhead{Preprint}

\definecolor{softgreen}{HTML}{E4F3E8}
\definecolor{rulegray}{HTML}{E6E6E6}
\definecolor{checkgreen}{HTML}{4F8A5B}
\definecolor{crossred}{HTML}{B46363}
\newcommand{\cmark}{{\color{checkgreen}\ding{51}}}
\newcommand{\xmark}{{\color{crossred}\ding{55}}}
\arrayrulecolor{rulegray}
\setlength{\arrayrulewidth}{0.35pt}
\renewcommand{\arraystretch}{1.22}

\begin{abstract}
Synthesizable molecular optimization seeks to improve target properties while ensuring that molecular modifications follow feasible synthetic pathways. Existing synthesis-aware methods typically rely on exploring a large space of candidate transformations defined by reaction templates and purchasable building blocks. This search becomes even more challenging when property improvement requires multiple reaction steps, as the space expands further along the pathway. To address this challenge, we introduce \textsc{MolReAct}, which reformulates molecular optimization as search over compact reaction spaces proposed by a tool-augmented large language model (LLM). At each step, the LLM combines its prior chemical knowledge with cheminformatics tools to identify a molecule-specific set of compatible reactions, preserving synthesizability while making multi-step optimization feasible. Given this compact action space, we further leverage Group Relative Policy Optimization (GRPO) with the terminal oracle reward to improve long-term decision-making over multiple reaction steps. Across diverse molecular optimization tasks, \textsc{MolReAct} achieves the highest Top-10 score on 11 of 14 tasks and the best sample efficiency on 12 of 14 tasks, outperforming existing baselines under limited oracle budgets. Beyond these gains, \textsc{MolReAct} also provides each optimized molecule with a template-grounded synthetic pathway.
\end{abstract}


\section{Introduction}

Molecular optimization is a central task in drug discovery that seeks to identify molecules with improved target properties~\citep{lead_survey}. However, high property scores alone are not enough for practical use, since optimized molecules must also be chemically valid and practically accessible through feasible synthetic transformations. Therefore, improving target properties while ensuring that molecules can be reliably synthesized remains a key research challenge~\citep{synthesis_survey}.


Prior work on synthesis-aware molecular optimization typically searches a combinatorial reaction space built from validated templates and predefined building blocks, using generative or search-based methods to construct feasible synthetic pathways~\citep{SyntheMol,TRACER,MCLM,VAE,Bayesian,MOBO,RGFN,SynFlowNet,RXNFLOW,MolSearch}. However, searching over numerous reaction templates and large building-block libraries yields an enormous number of candidate reactions. Identifying promising reactions among these candidates requires extensive oracle calls and substantial computational resources. This problem becomes even worse over multiple reaction steps, as the search space expands further at each step, making multi-step optimization more difficult in practice.

\begin{figure}[t]
\centering
\includegraphics[width=1.0\linewidth]{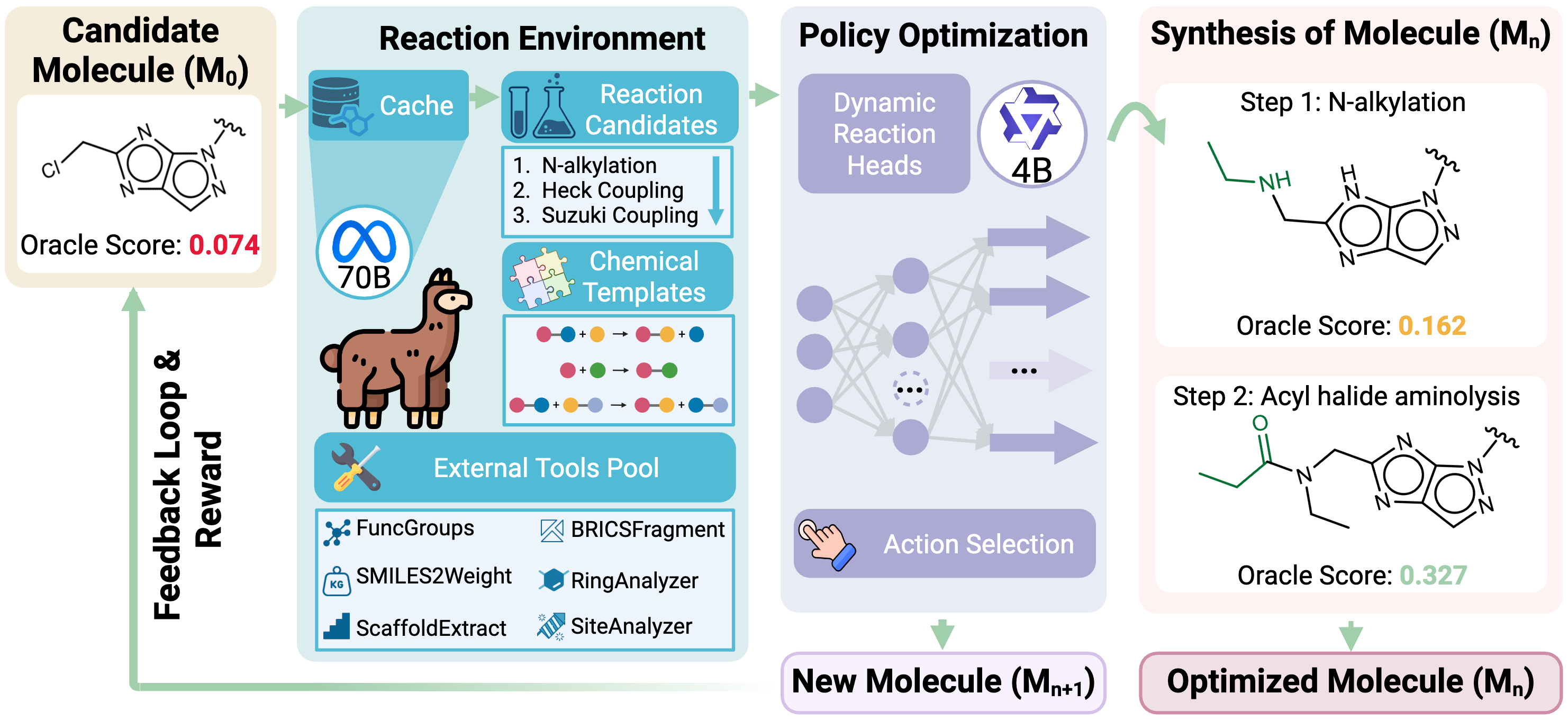}
\caption{
Overview of MolReAct. The reaction environment performs template matching and tool-augmented analysis to propose feasible candidates. The policy selects among candidates or stops the trajectory. The right panel shows a two-step optimization via N-alkylation and acyl halide aminolysis.
}

\label{fig:overview}
\end{figure}

To address this search-space bottleneck, we reformulate molecular optimization as search over compact reaction spaces proposed by a tool-augmented large language model (LLM). Pretrained on extensive scientific corpora, LLMs demonstrate strong capabilities in molecular understanding and reaction reasoning.~\citep{chemistry_llm1,chemistry_llm2,chemistry_llm3,chemistry_llm4,chemistry_llm5}. Tool augmentation can further strengthen this capability by incorporating explicit chemical and structural information from external chemistry tools~\citep{chemcrow,tool_based_chemistry_LLM1,tool_based_chemistry_LLM2}. This reformulation shifts the role of the LLM from complex optimized molecule generation to targeted search-space guidance, where at each step it dynamically constructs a compact reaction space tailored to the current molecule, making multi-step optimization feasible.

Building on this idea, we propose \textsc{MolReAct}, a synthesis-aware framework for multi-step molecular optimization from a given starting molecule. At each step, substructure matching identifies reaction templates compatible with the current molecule. The LLM then combines its chemical knowledge with specialized cheminformatics tools to select promising reactions. Instantiating them with purchasable building blocks from the Enamine catalog~\citep{enamine} yields a compact and synthesizable reaction space. However, reaching the target properties usually takes several reaction steps, so effective optimization requires planning across the entire pathway. We therefore deploy a policy model optimized with Group Relative Policy Optimization (GRPO), using the terminal oracle reward to evaluate the long-term impact of each reaction across the pathway for multi-step optimization. To keep this optimization efficient, we cache LLM proposals for previously seen molecules, avoiding repeated inference while maintaining a comparable computational cost. Across TDC benchmarks~\citep{tdc} and a proxy-based binding affinity task~\citep{sEH_bengio}, \textsc{MolReAct} achieves the highest Top-10 score on 11 out of 14 tasks and the best sample efficiency on 12 out of 14 tasks, outperforming existing baselines under limited oracle budgets. Further analysis shows that tool-guided proposals improve candidate quality at each step, which in turn enables more effective policy optimization across multiple steps. By constructing a dynamic molecule-specific action space, \textsc{MolReAct} provides each optimized molecule with a template-grounded pathway.
\section{Related Work}

\subsection{Synthesizable Molecular Optimization}
\label{sec:related_synth_opt}

Molecular optimization often prioritizes target properties without enforcing synthetic feasibility. Recent approaches address this through two main strategies. The first restricts the search space using predefined building block libraries and validated reaction templates. Algorithms like variational autoencoders~\citep{VAE}, Bayesian optimization~\citep{Bayesian}, Monte Carlo tree search~\citep{SyntheMol,TRACER,MolSearch,ClickGen}, and GFlowNets~\citep{MOBO,RGFN,SynFlowNet, RXNFLOW} navigate this space to construct valid reaction sequences. However, exploring these extensive reaction networks without directional guidance demands massive computational resources and heavy oracle evaluations. An alternative strategy projects given molecules into the synthesizable space to suggest valid analogs. These approaches generally learn this mapping by training separate projection models~\citep{Synformer, Reasyn, PSSM, anolog_paper}, often integrated with genetic algorithms or by fine-tuning large language models directly~\citep{MCLM,SynLlama}. Although these methods produce synthetically feasible analogs, it incurs additional training overhead. Rather than enumerating predefined reactions or training separate projection models, MolReAct performs molecule-conditioned optimization from a given starting molecule, using a tool-augmented LLM to dynamically construct a compact chemical action space explicitly tailored to its chemical context.

\subsection{Tool-Augmented LLM for Chemical Reasoning}
\label{sec:related_llm_search}

Large language models have shown strong capabilities in chemical reasoning and property interpretation~\citep{chemistry_llm1,chemistry_llm2,chemistry_llm3,chemistry_llm4,chemistry_llm5}. However, when applied to generative molecular design, their internal knowledge alone often yields chemically invalid or implausible structures~\citep{validity_smiles}. To address this limitation, recent works augment LLMs with external chemistry tools to ground the generation process in explicit physical rules, such as reaction planners, safety checkers, and molecular databases~\citep{chemcrow,tool_based_chemistry_LLM1,tool_based_chemistry_LLM2}. While these tool-augmented systems significantly improve the chemical validity of the outputs, they are primarily designed for executing predefined workflows, answering chemical queries, or performing single-step retrospective synthesis. In practice, these systems often struggle with long-term decision-making during multi-step optimization.
\section{Methodology}

\subsection{Problem Formulation}
\label{sec:problem_formulation}

We formulate synthesizable molecular optimization as a Markov Decision Process (MDP), $\mathcal{M} = (\mathcal{S}, \mathcal{A}, \mathcal{P}, \mathcal{R})$. At step $t$, the state $s_t \in \mathcal{S}$ is the chemically valid molecule. An action $a_t \in \mathcal{A}(s_t)$ either specifies a reaction through a template and purchasable building block or stops the optimization. The transition is deterministic, with $\mathcal{P}(s_{t+1}\mid s_t,a_t)$ determined by executing the selected reaction. An episode terminates when the policy selects the stop action $a_{\text{stop}}$, reaches the maximum depth $T_{\max}$, or no reaction template matches the current molecule. The oracle evaluates the final molecule $s_T$, giving the corresponding terminal reward $r_T = \mathrm{Oracle}(s_T)$. The policy $\pi_{\theta}(a_t \mid s_t)$ is then trained to maximize the expected terminal reward, $J(\pi_\theta) = \mathbb{E}{\pi\theta}[r_T]$. Since every reaction action comes from a validated template, each optimization trajectory defines an explicit template-grounded pathway.

\subsection{LLM Agent as a Dynamic Reaction Environment}
\label{sec:environment}

We construct an action space tailored to each molecule through a two-stage process. The first stage grounds all actions in validated reaction templates to enforce template validity. The second leverages LLM chemical knowledge to select relevant transformations and instantiate concrete building blocks.

\paragraph{Template Matching.}
Following~\citet{Synformer} and~\citet{Reasyn}, we adopt a curated set of 115 validated reaction templates that include uni-, bi-, and tri-molecular reactions. Given the current molecule $s_t$ represented as a SMILES string, we perform substructure matching against the reactant patterns in each template. Only templates whose reactant patterns are present in $s_t$ are retained, yielding a matched subset $\mathcal{T}_{\mathrm{match}}(s_t) \subseteq \mathcal{T}$. If no template matches, the trajectory terminates early.

\paragraph{LLM-Guided Action Space.}
The agent receives a task-specific natural-language description of the optimization objective (Appendix Table~\ref{tab:property_descriptions}). Given $\mathcal{T}_{\mathrm{match}}(s_t)$, the tool-augmented LLM analyzes the current molecule, selects the most chemically relevant templates, and proposes candidate reactions by specifying the building block required for each selected template. The agent operates within the ReAct framework~\citep{react}, interleaving reasoning with invocations of specialized chemical analysis tools (Table~\ref{tab:tools_extended}). These tools provide information about functional groups, scaffolds, ring systems, BRICS fragments, reactive sites, and molecular weight, enabling the agent to understand the molecular structure, select appropriate templates, and propose chemically grounded building blocks.

\newpage

Before execution, we canonicalize each additional reactant specified by the LLM and retain a proposal only if every required building block exactly matches an entry in the Enamine catalog of building blocks~\citep{enamine}. Reactions without an additional reactant are retained. The remaining proposals are executed via their templates to produce candidate products, and proposals that fail execution or yield invalid products are discarded. The resulting filtered action space at state $s_t$ is
\begin{equation}
    \mathcal{A}(s_t) = \left\{a \in \mathrm{LLM}_\phi\bigl(\mathrm{Tools}(s_t),\; \mathcal{T}_{\mathrm{match}}(s_t),\; \mathrm{obj}\bigr) : \mathrm{Catalog}(a) \land \mathrm{Valid}(a)\right\} \cup \{a_{\mathrm{stop}}\},
\end{equation}
where $\mathrm{Catalog}(a)$ denotes the Enamine match for all required additional reactants and $\mathrm{Valid}(a)$ denotes successful template execution with a valid product. The $\mathrm{obj}$ defines the specific task objective. The stop action $a_{\mathrm{stop}}$ immediately terminates the trajectory. Proposed reactions and the stop action jointly define a dynamic action space tailored to the current molecule at each optimization step.

\paragraph{Reaction Caching.}
To reduce redundant LLM invocations during RL exploration, we cache the action space $\mathcal{A}(s_t)$ and transition outcomes for each previously encountered molecule, keyed by SMILES within each task. During training, the environment queries this cache before invoking the LLM agent, reusing the same action space and transition outcomes after the first query. On a cache hit, reactions and pre-computed products are retrieved instantly. This significantly accelerates GRPO training by amortizing the LLM inference cost across repeated visits to the same molecular states.

\subsection{Trajectory Optimization via Reinforcement Learning}
\label{sec:rl_grpo}

Given the synthesis-constrained action space from Section~\ref{sec:environment}, We further train a policy model to select candidate reactions over multi-step trajectories, maximizing the terminal oracle reward.

\paragraph{Policy Architecture.}
The policy model is a Qwen-3-4B language model with a linear action head. At each step $t$, it reads a single prompt containing the current molecule SMILES $s_t$ and all candidate reactions, each numbered and described by its template name, reactant SMILES, and product SMILES. Encoding the molecule and every candidate together in one forward pass lets self-attention compare them directly and link each number to the reaction it labels. The action head then reads the final hidden state at each numbered candidate position and outputs a single logit for it. Through this process, the policy learns to dynamically score the proposed reactions. Unused slots are masked with $-\infty$ before the softmax, giving a distribution $\pi_\theta(a_t \mid s_t)$ over the current valid actions.

\paragraph{Group Relative Policy Optimization.}
For each initial molecule, the policy samples a group of $G$ independent trajectories. Each trajectory receives a single terminal reward $r_T = \mathrm{Oracle}(s_T)$, which is standardized within the group by subtracting the group mean and dividing by the group standard deviation to obtain a group-relative advantage. This shared advantage is then assigned to every step in the corresponding trajectory, enabling trajectory-level credit assignment. The policy is then updated by maximizing the standard clipped surrogate policy objective with additional KL regularization
\begin{equation}
\mathcal{L}_{\text{GRPO}}(\theta) = \mathbb{E} \left[ \min\!\left( \rho_t(\theta) A, \mathrm{clip}(\rho_t(\theta), 1 - \epsilon, 1 + \epsilon) A \right) \right] - \beta \, \mathbb{D}_{\text{KL}}\!\left( \pi_\theta(\cdot \mid s_t) \;\|\; \pi_{\mathrm{ref}}(\cdot \mid s_t) \right),
\end{equation}
where $\rho_t(\theta) = \frac{\pi_\theta(a_t \mid s_t)}{\pi_{\theta_{\text{old}}}(a_t \mid s_t)}$ is the importance sampling ratio with $\pi_{\theta_{\text{old}}}$ the behavior policy at rollout, $\pi_{\mathrm{ref}}$ the fixed initial policy used as a KL anchor, $\epsilon$ the clipping threshold, and $\beta$ the KL penalty strength preventing catastrophic deviation from $\pi_{\mathrm{ref}}$. An entropy bonus on $\pi_\theta$ is additionally applied during training to encourage exploration (Appendix~\ref{app:implementation}). The KL divergence is computed over valid actions at each state, naturally accommodating variable-size action spaces across different states.

\section{Experiments}

\subsection{Experimental Setup}
\label{sec:exp_settings}

MolReAct is evaluated on 14 molecular optimization tasks against both reaction-driven and LLM-based baselines under a fixed oracle budget. All methods share the same held-out starting molecules and reaction template library, so performance differences primarily reflect their search strategies.

\newpage

\newcommand{\sd}[1]{\raisebox{-0.3ex}{\fontsize{4}{5}\selectfont$\pm$#1}}
\newcommand{\gp}[1]{\,\raisebox{-0.25ex}{\fontsize{5}{6}\selectfont(+#1)}}
\definecolor{mctspurple}{HTML}{8172B3}

\begin{center}
\centering
\scriptsize
\captionof{table}{Top-10 performance on 14 molecular optimization tasks. Best in \textbf{bold}, second-best \underline{underlined}. $\checkmark$: synthesis-constrained methods. All methods report mean$\pm$sd over 3 runs.}
\label{tab:top10}
\setlength{\tabcolsep}{3.8pt}
\renewcommand{\arraystretch}{1.2}
\resizebox{\textwidth}{!}{%
\begin{tabular}{lcccccccc}
\toprule
& \textit{Initial} & Graph GA & ReaSyn & SynFormer & DrugAssist & LDMOL & mCLM & \textbf{MolReAct} \\
Synthesizable & -- & $\times$ & $\checkmark$ & $\checkmark$ & $\times$ & $\times$ & $\checkmark$ & $\checkmark$ \\
\midrule
amlodipine\_mpo       & 0.389 & \textbf{0.563}\sd{0.076} & 0.438\sd{0.005} & 0.452\sd{0.006} & 0.428\sd{0.008} & 0.402\sd{0.008} & 0.387\sd{0.002} & \underline{0.498}\sd{0.013} \\
celecoxib\_rediscovery & 0.182 & 0.351\sd{0.247} & 0.224\sd{0.005} & \underline{0.370}\sd{0.005} & 0.200\sd{0.003} & 0.198\sd{0.007} & 0.213\sd{0.004} & \textbf{0.393}\sd{0.021} \\
DRD2                   & 0.599 & 0.808\sd{0.024} & \textbf{0.996}\sd{0.003} & \textbf{0.996}\sd{0.004} & 0.679\sd{0.006} & 0.651\sd{0.009} & 0.236\sd{0.007} & \underline{0.987}\sd{0.009} \\
fexofenadine\_mpo     & 0.427 & 0.713\sd{0.035} & 0.684\sd{0.005} & \underline{0.760}\sd{0.075} & 0.589\sd{0.002} & 0.535\sd{0.005} & 0.545\sd{0.006} & \textbf{0.787}\sd{0.031} \\
GSK3\_beta            & 0.397 & \underline{0.673}\sd{0.035} & 0.660\sd{0.005} & 0.610\sd{0.005} & 0.319\sd{0.045} & 0.396\sd{0.008} & 0.173\sd{0.006} & \textbf{0.722}\sd{0.028} \\
JNK3                   & 0.269 & 0.422\sd{0.178} & \underline{0.470}\sd{0.003} & \underline{0.470}\sd{0.006} & 0.243\sd{0.003} & 0.283\sd{0.010} & 0.078\sd{0.002} & \textbf{0.541}\sd{0.038} \\
median\_1              & 0.100 & 0.214\sd{0.028} & 0.175\sd{0.008} & \underline{0.234}\sd{0.003} & 0.102\sd{0.005} & 0.108\sd{0.008} & 0.125\sd{0.008} & \textbf{0.253}\sd{0.013} \\
median\_2              & 0.126 & 0.141\sd{0.003} & \underline{0.166}\sd{0.002} & 0.151\sd{0.007} & 0.135\sd{0.002} & 0.139\sd{0.009} & 0.141\sd{0.009} & \textbf{0.186}\sd{0.012} \\
osimertinib\_mpo      & 0.306 & 0.743\sd{0.011} & 0.731\sd{0.009} & \underline{0.819}\sd{0.008} & 0.731\sd{0.002} & 0.553\sd{0.006} & 0.703\sd{0.002} & \textbf{0.825}\sd{0.020} \\
perindopril\_mpo      & 0.305 & 0.435\sd{0.006} & 0.381\sd{0.008} & \underline{0.460}\sd{0.009} & 0.363\sd{0.006} & 0.321\sd{0.002} & 0.360\sd{0.004} & \textbf{0.487}\sd{0.014} \\
ranolazine\_mpo       & 0.081 & 0.645\sd{0.034} & 0.657\sd{0.012} & \underline{0.795}\sd{0.011} & 0.458\sd{0.012} & 0.170\sd{0.012} & 0.195\sd{0.011} & \textbf{0.812}\sd{0.016} \\
sEH                    & 0.716 & 0.909\sd{0.005} & \textbf{0.976}\sd{0.005} & 0.918\sd{0.058} & 0.652\sd{0.030} & 0.724\sd{0.007} & 0.654\sd{0.006} & \underline{0.934}\sd{0.013} \\
sitagliptin\_mpo      & 0.007 & \underline{0.329}\sd{0.070} & 0.283\sd{0.012} & 0.295\sd{0.024} & 0.002\sd{0.002} & 0.014\sd{0.006} & 0.044\sd{0.004} & \textbf{0.391}\sd{0.017} \\
zaleplon\_mpo         & 0.133 & 0.393\sd{0.023} & 0.326\sd{0.006} & \underline{0.429}\sd{0.010} & 0.236\sd{0.011} & 0.217\sd{0.005} & 0.347\sd{0.008} & \textbf{0.476}\sd{0.016} \\
\midrule
Average score              & 0.288 & 0.524\sd{0.035} & 0.512\sd{0.003} & \underline{0.554}\sd{0.005} & 0.367\sd{0.009} & 0.337\sd{0.006} & 0.300\sd{0.006} & \textbf{0.592}\sd{0.014} \\
Rank (Gain \%) & -- & 3\gp{82} & 4\gp{78} & \underline{2}\gp{92} & 5\gp{27} & 6\gp{17} & 7\gp{4} & \textbf{1}\gp{106} \\
\bottomrule
\end{tabular}%
}
\end{center}

\begin{wraptable}[20]{r}{0.55\textwidth}
\vspace{-1\baselineskip}
\centering
\caption{AUC$_{10}$ sample efficiency comparison across all 14 optimization tasks. Best in \textbf{bold}, second-best \underline{underlined}. All methods report mean$\pm$sd over 3 runs.}
\label{tab:auc10}
\scriptsize
\setlength{\tabcolsep}{3pt}
\renewcommand{\arraystretch}{1.2}
\resizebox{\linewidth}{!}{%
  \begin{tabular}{lcccc}
    \toprule
    Property & Graph GA & ReaSyn & SynFormer & \textbf{MolReAct} \\
    \midrule
    amlodipine\_mpo        & 0.423\sd{0.050} & 0.432\sd{0.015} & \underline{0.445}\sd{0.011} & \textbf{0.471}\sd{0.017} \\
    celecoxib\_rediscovery & 0.312\sd{0.165} & 0.222\sd{0.006} & \underline{0.344}\sd{0.021} & \textbf{0.387}\sd{0.012} \\
    DRD2                   & 0.781\sd{0.114} & \underline{0.989}\sd{0.010} & \textbf{0.993}\sd{0.005} & 0.980\sd{0.007} \\
    fexofenadine\_mpo      & 0.693\sd{0.002} & 0.669\sd{0.008} & \underline{0.716}\sd{0.059} & \textbf{0.732}\sd{0.019} \\
    GSK3\_beta             & 0.598\sd{0.012} & \underline{0.649}\sd{0.107} & 0.605\sd{0.093} & \textbf{0.684}\sd{0.024} \\
    JNK3                   & 0.418\sd{0.009} & 0.454\sd{0.122} & \underline{0.466}\sd{0.091} & \textbf{0.524}\sd{0.031} \\
    median\_1              & 0.189\sd{0.006} & 0.172\sd{0.005} & \underline{0.221}\sd{0.026} & \textbf{0.231}\sd{0.023} \\
    median\_2              & 0.125\sd{0.005} & \underline{0.164}\sd{0.017} & 0.149\sd{0.004} & \textbf{0.171}\sd{0.011} \\
    osimertinib\_mpo       & 0.719\sd{0.004} & 0.726\sd{0.021} & \underline{0.791}\sd{0.011} & \textbf{0.810}\sd{0.010} \\
    perindopril\_mpo       & 0.398\sd{0.011} & 0.378\sd{0.016} & \underline{0.439}\sd{0.010} & \textbf{0.475}\sd{0.015} \\
    ranolazine\_mpo        & 0.612\sd{0.002} & 0.606\sd{0.030} & \underline{0.746}\sd{0.010} & \textbf{0.781}\sd{0.016} \\
    sEH                    & 0.877\sd{0.031} & \textbf{0.915}\sd{0.043} & 0.909\sd{0.030} & \underline{0.910}\sd{0.014} \\
    sitagliptin\_mpo       & \underline{0.310}\sd{0.014} & 0.281\sd{0.086} & 0.281\sd{0.054} & \textbf{0.355}\sd{0.011} \\
    zaleplon\_mpo          & 0.315\sd{0.007} & 0.322\sd{0.042} & \underline{0.410}\sd{0.022} & \textbf{0.447}\sd{0.022} \\
    \midrule
    Average score          & 0.484\sd{0.010} & 0.499\sd{0.025} & \underline{0.537}\sd{0.009} & \textbf{0.568}\sd{0.008} \\
    \bottomrule
  \end{tabular}%
}
\end{wraptable}

\paragraph{Tasks and Evaluation Metrics.}
Following \citet{Synformer} and \citet{Reasyn}, we evaluate MolReAct on 13 property optimization tasks from the Therapeutic Data Commons (TDC)~\citep{tdc} and one proxy-based binding affinity task for soluble Epoxide Hydrolase (sEH)~\citep{sEH_bengio}. The 14 tasks cover multi-parameter optimization (MPO), molecular rediscovery and median objectives, and protein-target binding activity prediction. Each task is allowed 10,000 oracle calls under a controlled-start protocol for molecule-conditioned optimization (Appendix~\ref{app:oracle_budget}). All methods begin from the same 100 held-out molecules, ensuring a fair comparison. All synthesizable methods share the same 115-template reaction library~\citep{Reasyn}. We report the \textit{Top-10 score} over all evaluated molecules and compute its Area Under the Curve (AUC$_{10}$) as a function of oracle calls to evaluate sample efficiency. We also report the Top-10 score of the starting molecules and each method's relative gain over it for comparison.

\paragraph{Dataset and Starting Molecule Selection.}
Each task starts from a given molecule selected from ZINC-250K~\citep{zinc} using the filtering protocol of \citet{MolSearch}. We retain molecules with QED above 0.6 to ensure drug-likeness. For protein-target activity tasks (JNK3, DRD2, GSK3$\beta$, and sEH), starting molecules are restricted to oracle scores between 0.1 and 0.8, while for the remaining tasks, they are selected from the 60th--80th percentile of the corresponding property distribution. These criteria maintain initial molecular quality while preserving the potential for meaningful property improvement. From the filtered pool, we randomly sample 1,000 molecules for policy training and 100 held-out molecules as starting molecules for evaluation across all methods.

\paragraph{Model and Training Configuration.}
Our framework involves two language models serving distinct roles. A Llama-3.3-70B model~\citep{llama}, deployed via the vLLM engine~\citep{vllm}, operates within the ReAct framework as the environment agent, leveraging chemical tools to analyze molecular structures and propose template-grounded candidate reactions at each step. We retain those proposals with building blocks available in the Enamine catalog~\citep{enamine} before reaction execution. The policy agent, a Qwen-3-4B model~\citep{qwen3}, is the sole trainable component, optimized via GRPO to learn reaction selection strategies that maximize long-term reward across multi-step trajectories. During training, we impose a maximum depth of 5 steps to prioritize synthetic efficiency and reduce side reactions from excessively long pathways.

\paragraph{Baselines.}
We compare against two categories of baselines. The first includes reaction-driven methods such as \textit{GraphGA}~\citep{GraphGA}, \textit{ReaSyn}~\citep{Reasyn}, and \textit{SynFormer}~\citep{Synformer}. \textit{ReaSyn} and \textit{SynFormer} project a molecule into synthesizable space to suggest analogs. The second category includes LLM-based frameworks such as \textit{DrugAssist}~\citep{DrugAssist} and \textit{LDMol}~\citep{LDMol}, which generate modified molecules from natural language instructions, and \textit{mCLM}~\citep{MCLM}, which further prioritizes synthesizability. These LLM-based baselines use the same task-specific objective descriptions and initial molecules as MolReAct.

\subsection{Results and Analysis}
\label{sec:results}

\subsubsection{Comparative Performance}
\label{sec:main_results}

\begin{wrapfigure}[28]{r}{0.50\textwidth}
\vspace{-1\baselineskip}
\centering
\captionof{table}{Molecular quality of the top-100 optimized molecules, averaged over the 14 tasks (mean$\pm$sd). Best in \textbf{bold}, second-best \underline{underlined}.}
\label{tab:quality}
\scriptsize
\setlength{\tabcolsep}{2.5pt}
\renewcommand{\arraystretch}{1.2}
\resizebox{\linewidth}{!}{%
  \begin{tabular}{lccccc}
    \toprule
    Method & IntDiv$\uparrow$ & SA$\downarrow$ & AiZynth$\uparrow$ & QED$\uparrow$ & MW$\downarrow$ \\
    \midrule
    \textit{Initial} & 0.870\sd{0.007} & 2.98\sd{0.21} & 0.574\sd{0.107} & 0.782\sd{0.020} & 333\sd{8.1} \\
    Graph GA          & 0.518\sd{0.037} & 3.24\sd{0.21} & 0.255\sd{0.180} & 0.525\sd{0.029} & 376\sd{22.7} \\
    ReaSyn            & 0.576\sd{0.024} & \textbf{2.81}\sd{0.14} & 0.568\sd{0.331} & 0.668\sd{0.046} & 347\sd{15.4} \\
    SynFormer         & 0.631\sd{0.045} & \underline{2.84}\sd{0.19} & \textbf{0.612}\sd{0.287} & 0.636\sd{0.022} & 393\sd{28.9} \\
    DrugAssist        & 0.873\sd{0.018} & 3.26\sd{0.25} & 0.433\sd{0.090} & 0.597\sd{0.038} & 389\sd{18.1} \\
    LDMOL             & \underline{0.875}\sd{0.032} & 3.08\sd{0.15} & 0.506\sd{0.083} & \textbf{0.744}\sd{0.051} & 340\sd{25.6} \\
    mCLM              & 0.848\sd{0.049} & 3.29\sd{0.23} & 0.516\sd{0.080} & 0.635\sd{0.031} & \underline{325}\sd{12.8} \\
    \textbf{MolReAct} & \textbf{0.881}\sd{0.027} & 2.94\sd{0.17} & \underline{0.581}\sd{0.073} & \underline{0.673}\sd{0.043} & \textbf{321}\sd{30.3} \\
    \bottomrule
  \end{tabular}%
}

\bigskip

\begin{minipage}[c]{0.49\linewidth}
  \centering
  \includegraphics[width=\linewidth]{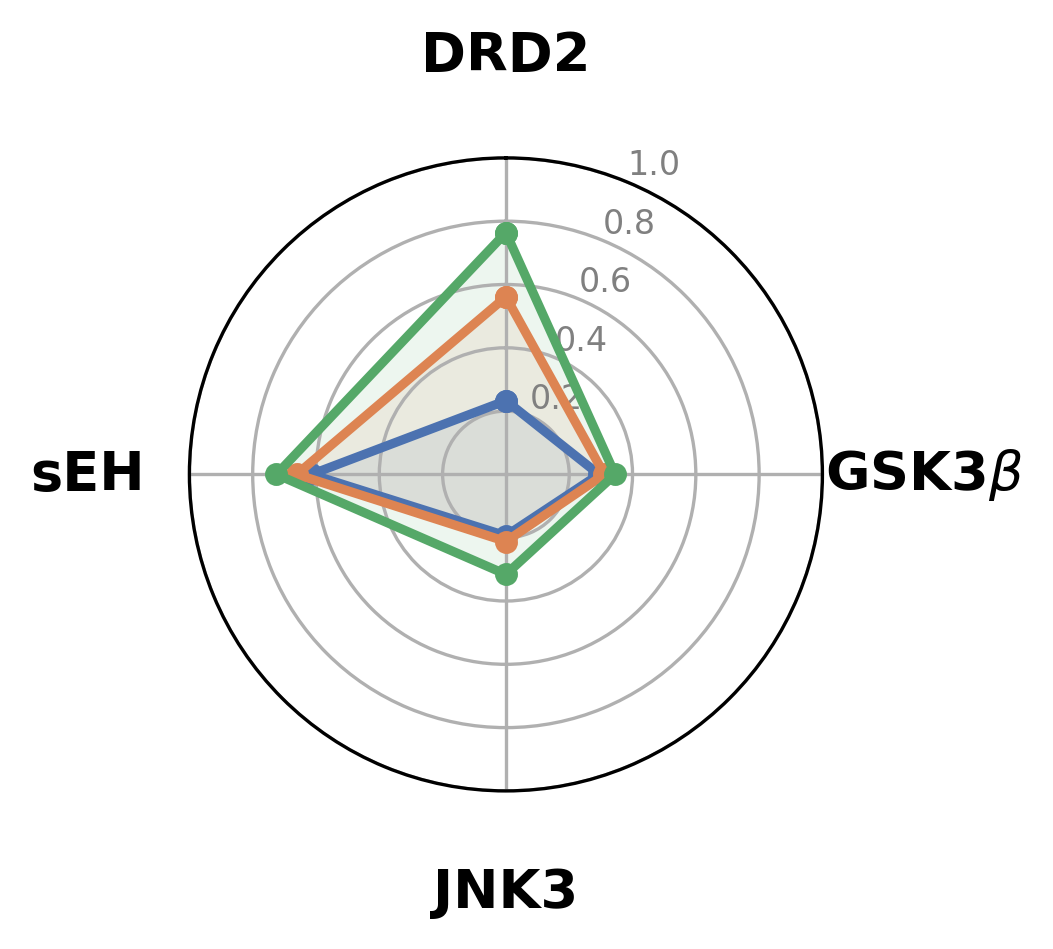}
  {\fontsize{8pt}{7pt}\selectfont
  (a) Reaction proposal:
  \textcolor{plotblue}{Random Template},
  \textcolor{plotorange}{Vanilla LLM},
  \textcolor{plotgreen}{Tool-LLM}}
\end{minipage}%
\hspace{1pt}%
\begin{minipage}[c]{0.49\linewidth}
  \centering
  \includegraphics[width=\linewidth]{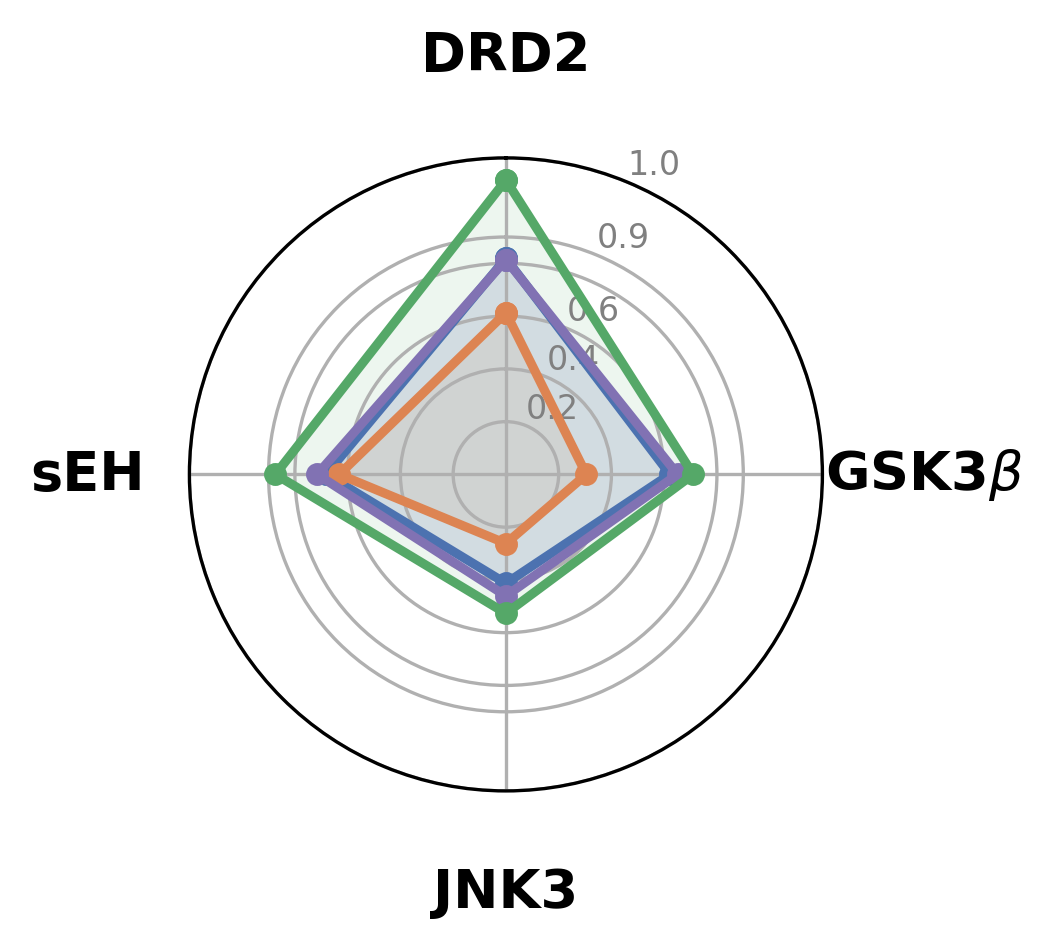}
  {\fontsize{8pt}{7pt}\selectfont
  (b) Multi-step optimization:
  \textcolor{plotblue}{Greedy Search},
  \textcolor{plotorange}{Random Policy},
  \textcolor{mctspurple}{MCTS},
  \textcolor{plotgreen}{MolReAct}}
\end{minipage}
\captionof{figure}{Ablation on tool-guided proposal and policy optimization across target activity tasks.}
\label{fig:radar_ablation}
\end{wrapfigure}

Table~\ref{tab:top10} reports the Top-10 scores across all 14 tasks. MolReAct achieves the highest average score of 0.592 and the largest relative gain over the starting Top-10 scores at 106\%. It performs best on 11 tasks and significantly outperforms all six baselines under paired Wilcoxon signed-rank tests (Appendix~\ref{app:significance}, $p<0.05$). Table~\ref{tab:auc10} reports AUC$_{10}$, which measures optimization quality over cumulative oracle calls. MolReAct achieves the best sample efficiency on 12 tasks, with an overall average AUC$_{10}$ of 0.568 compared with 0.537 for SynFormer. It significantly outperforms all three baselines, indicating that MolReAct consistently finds high-scoring molecules with fewer oracle calls. This improvement extends across most optimization tasks.


Table~\ref{tab:quality} evaluates the overall quality of the top-100 optimized molecules from each task in terms of internal diversity (IntDiv), synthetic accessibility (SA), AiZynthFinder retrosynthesis success rate (AiZynth), drug-likeness (QED), and molecular weight (MW). MolReAct achieves the highest IntDiv of 0.881 and the lowest MW of 321 among all methods, while closely maintaining SA and AiZynth at levels comparable to the starting molecules across tasks. Although its QED decreases slightly, it still remains competitive with the other methods. Overall, these results indicate that MolReAct consistently improves the target properties across tasks without relying on molecular-size inflation or substantially compromising diversity and synthetic accessibility.

\subsubsection{Ablation Study}
\label{sec:ablation}

We conduct a progressive ablation to isolate the individual contributions of tool augmentation and RL-based policy optimization. We focus on the four protein-target activity tasks (JNK3, DRD2, GSK3$\beta$, sEH), as their well-defined biological targets and continuous activity scores provide the clearest and most interpretable signal for distinguishing the effect of each component. Results are visualized in Figure~\ref{fig:radar_ablation}, with exact numerical values for all variants provided in Table~\ref{tab:ablation_numbers} in the appendix.

\paragraph{Tool-Guided Reaction Proposal.}
We first evaluate the quality of single-step reaction proposals. For each starting molecule, the LLM proposes candidate reactions, and their products are then evaluated by the oracle. We report the average Top-10 score to assess proposal quality. Three different proposal variants are compared.
\textbf{(i)~Random Template}, which randomly samples structurally matched reaction templates and compatible building blocks;
\textbf{(ii)~Vanilla LLM}, which selects reactions based on the model's internal chemical knowledge for the target property; and
\textbf{(iii)~Tool LLM}, which equips the LLM with domain-specific analytical tools (\emph{e.g.}, reactive site analysis, scaffold extraction) together with task-specific objective descriptions.
Figure~\ref{fig:radar_ablation}(a) shows that Vanilla LLM outperforms Random Template, while tool augmentation provides further gains and leads to the strongest overall performance for Tool LLM. We additionally run an \textbf{anonymized Tool LLM} that does not receive the objective description, which performs comparably to the Tool LLM across all evaluated tasks (Appendix Section~\ref{app:task_information}), indicating that proposal quality is driven primarily by the LLM's internal chemical knowledge and analytical tools rather than by any potential leakage from task information.

\paragraph{Policy-Guided Multi-Step Optimization.}
We then evaluate optimization strategies over multi-step trajectories. Fixing the reaction proposer as Tool LLM, we compare four policies:
\textbf{(i)~Random Policy}, which randomly selects a reaction from the proposed candidates at each step;
\textbf{(ii)~Greedy Search}, which selects the candidate with the largest local improvement;
\textbf{(iii)~Monte Carlo Tree Search}, which plans over the action space using UCT (Appendix~\ref{app:mcts}); and
\textbf{(iv)~MolReAct (GRPO)}, the full framework with a trained policy model.
Figure~\ref{fig:radar_ablation}(b) shows that Greedy Search outperforms Random Policy, while tending to get trapped in local optima. MCTS improves over Greedy Search on most tasks through multi-step planning. MolReAct performs best overall. These results show that the learned policy provides additional benefits beyond the action space by enabling long-term decision-making. To examine whether these gains rely on task-specific information, we additionally train and evaluate the full framework using the \textbf{anonymized MolReAct}, which achieves performance comparable to MolReAct (Appendix Section~\ref{app:task_information}). This indicates that LLM guidance can provide an effective action space for multi-step policy optimization without explicit task-specific information.

\begin{wrapfigure}[22]{r}{0.41\textwidth} \centering \vspace{-1.5\baselineskip} \begin{subfigure}[b]{0.40\textwidth} \centering \includegraphics[width=\textwidth]{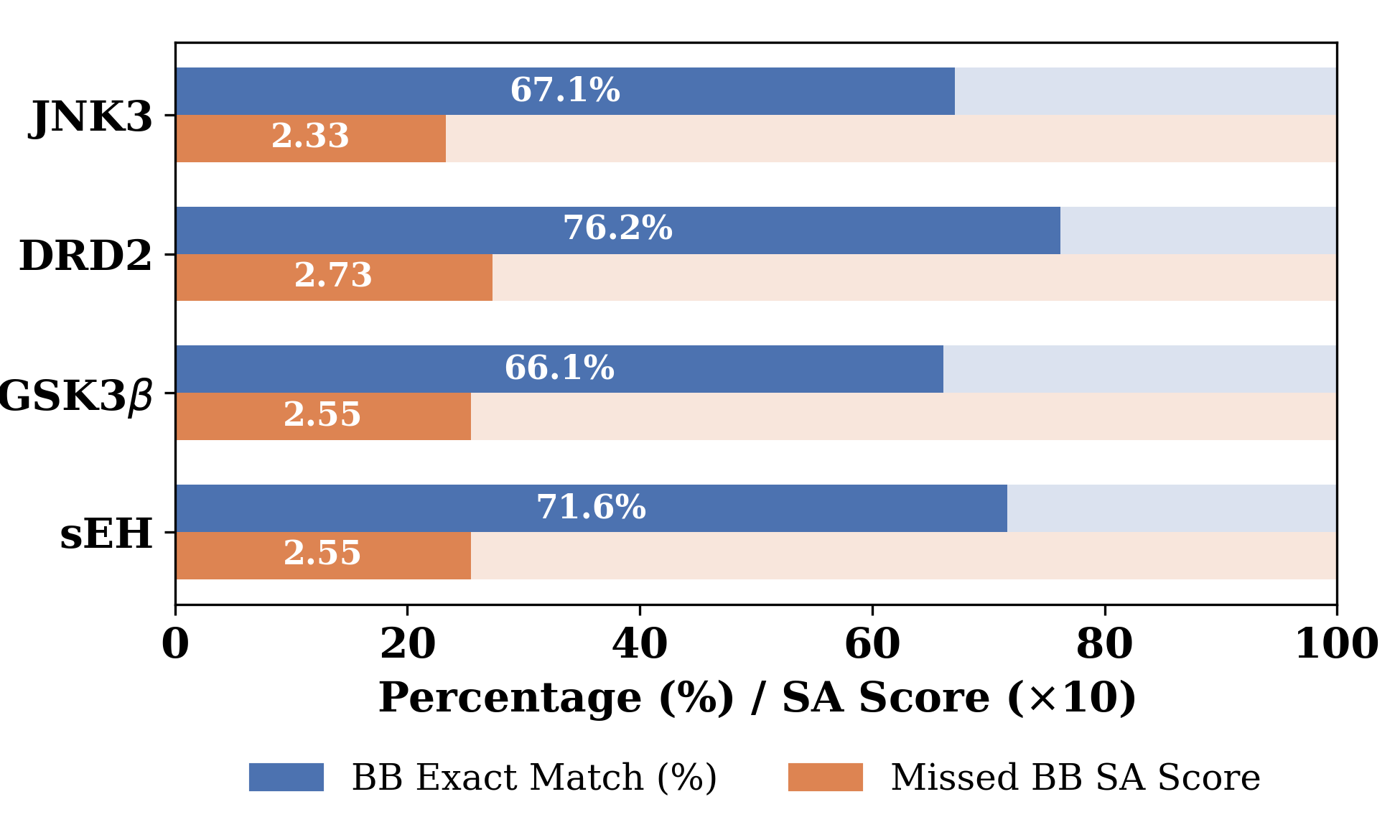} \caption{Availability before catalog filtering.} \label{fig:bb_availability} \end{subfigure} \begin{subfigure}[b]{0.40\textwidth} \centering \includegraphics[width=\textwidth]{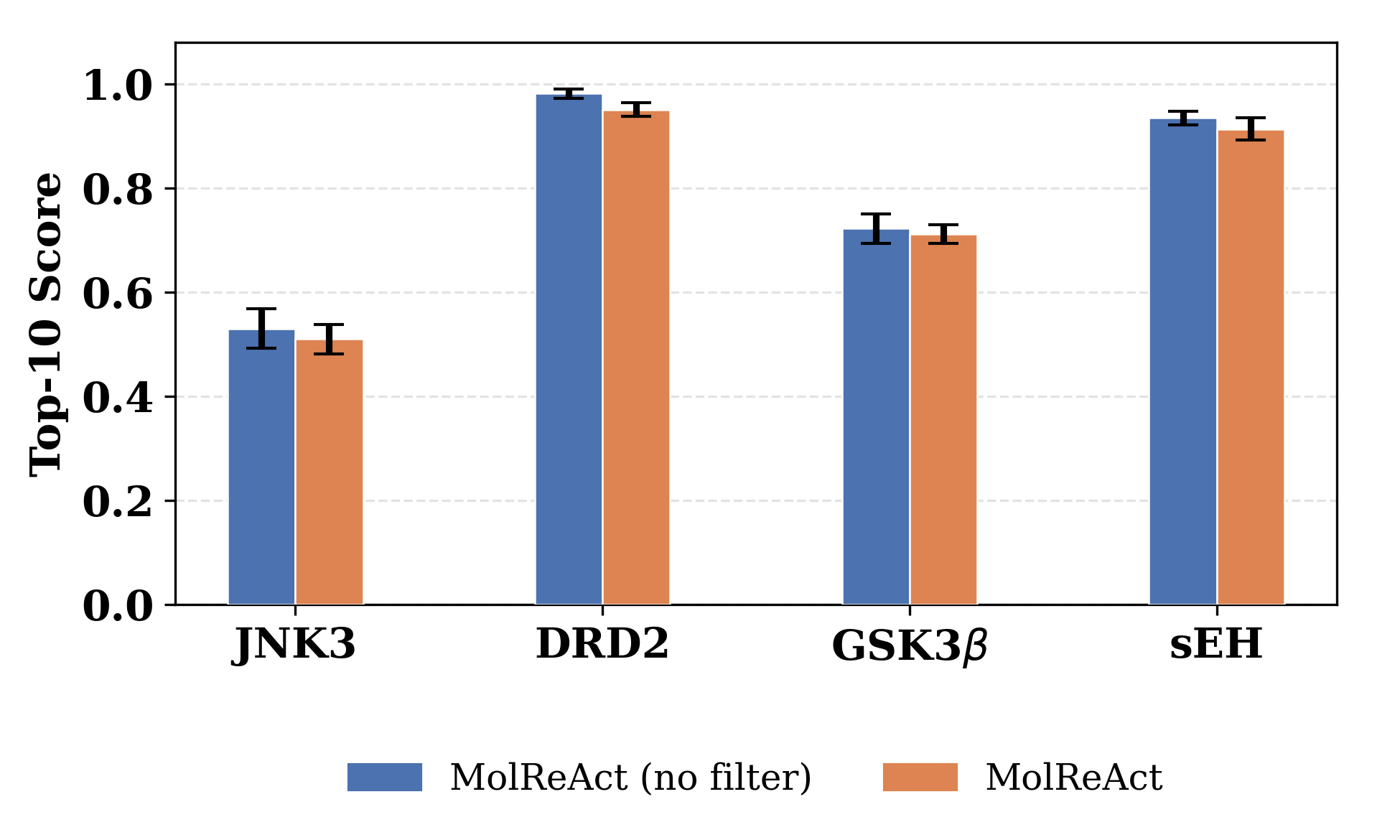} \caption{Top-10 with/without catalog filtering.} \label{fig:bb_quality} \end{subfigure} \caption{Effect of commercial filtering.} 
\label{fig:bb_filter} \end{wrapfigure}

\subsubsection{Commercial Building-Block Filtering}
\label{sec:bb_availability}

MolReAct restricts its action space to purchasable building blocks throughout training and evaluation. A building block is retained only if its canonical SMILES exactly matches an entry in the Enamine catalog~\citep{enamine}. This constraint is not overly restrictive in practice, as the LLM already proposes mostly purchasable reagents. As shown in Figure~\ref{fig:bb_availability}, $66.1$--$76.2\%$ of the unique proposed building blocks are directly available in the catalog. The unmatched candidate building blocks also remain synthetically accessible, with SA scores ranging from $2.33$ to $2.73$ (all below 3). Because exact matching does not capture close analogs, salt forms, or availability from other suppliers, the actual availability is likely higher.

To quantify the performance cost introduced by this constraint, we compare MolReAct with an otherwise identical variant trained without the purchasability filter on the four protein-target tasks. Enforcing purchasability throughout training and evaluation changes the overall Top-10 score by only $1.4$--$3.8\%$ (Figure~\ref{fig:bb_quality}), which falls within observed run-to-run variation on three of the four targets. Thus, restricting the action space to purchasable building blocks preserves nearly all overall optimization performance while ensuring that every selected reaction uses a commercially available reagent.

\subsubsection{Component Sensitivity Study}
\label{sec:llm_sensitivity}

MolReAct contains two components, the environment LLM and policy model. We vary each while holding the other fixed, asking how much capacity each requires on four protein-target activity tasks.

\paragraph{Environment LLM.}
Since the environment LLM defines the action space, we first examine how model scale affects overall single-step proposal quality. Using the same reaction-proposal protocol as Section~\ref{sec:ablation} on the four protein-target tasks, we compare four Llama model variants and report the Top-10 product score, coverage (the fraction of molecules with at least one valid reaction), and amortized GPU-seconds per molecule (Table~\ref{tab:llm_sensitivity}). Both Top-10 and coverage improve substantially on average from 3B to 70B, with coverage more than doubling from 34\% to 79\%. The 405B variant yields only a modest Top-10 improvement, while coverage decreases slightly, suggesting that the larger model is more selective and more likely to refrain when it cannot identify a confident transformation. Considering its roughly $6\times$ higher computational cost in practice, these results make 70B the best overall trade-off for our framework and motivate our choice of environment LLM.

\begin{wraptable}[18]{r}{0.35\textwidth}
\centering
\vspace{-1.5\baselineskip}

\begin{subtable}{\linewidth}
\centering
\caption{Environment LLM.}
\label{tab:llm_sensitivity}
\small
\setlength{\tabcolsep}{4pt}
\renewcommand{\arraystretch}{1.05}
\resizebox{\linewidth}{!}{%
\begin{tabular}{lccc}
\toprule
Environment LLM & Top-10 & Cov. & GPU-s \\
\midrule
Llama-3.2-3B          & 0.488 & 34\% & 6 \\
Llama-3.1-8B          & 0.515 & 35\% & 21 \\
\textbf{Llama-3.3-70B}         & \textbf{0.538} & \textbf{79\%} & \textbf{201} \\
Llama-3.1-405B & 0.575 & 74\% & 1238 \\
\bottomrule
\end{tabular}}
\end{subtable}

\vspace{4pt}

\begin{subtable}{\linewidth}
\centering
\caption{Policy model.}
\label{tab:policy_scale}
\small
\setlength{\tabcolsep}{4pt}
\renewcommand{\arraystretch}{1.05}
\resizebox{\linewidth}{!}{%
\begin{tabular}{lccc}
\toprule
Policy model & Top-10 & AUC$_{10}$ & Time (h) \\
\midrule
ECFP4+MLP & 0.640 & 0.600 & 9 \\
Qwen3-0.6B & 0.618 & 0.605 & 13 \\
Qwen3-1.7B & 0.721 & 0.712 & 25 \\
\textbf{Qwen3-4B}   & \textbf{0.796} & \textbf{0.775} & \textbf{40} \\
Qwen3-8B   & 0.798 & 0.753 & 86 \\
\bottomrule
\end{tabular}}
\end{subtable}

\caption{Sensitivity analysis for (a) LLM scale and (b) policy model. Bold denotes the selected models.}
\label{tab:llm_sensitivity_all}
\end{wraptable}

\paragraph{Policy Model.}
We next study the contribution of the learned policy model by fixing the environment agent (Tool-LLM) and varying the model used to select among candidate reactions (Table~\ref{tab:policy_scale}). To assess the benefit of using a language model as the policy, we replace it with a 2.29M-parameter MLP over ECFP4 fingerprints and template identity. The MLP is trained under the same GRPO objective and oracle budget (Appendix~\ref{app:ecfp_policy}). The ECFP4+MLP policy performs substantially worse than Qwen3-4B in both Top-10 and AUC$_{10}$, indicating that pretrained chemical representations improve effective downstream reaction selection and sample efficiency under the same action space. We then vary the size of the language-model policy. Performance improves steadily from Qwen3-0.6B to Qwen3-4B and then largely saturates across tasks, with Qwen3-8B providing little additional gain while more than doubling the training time. These results suggest that the policy benefits from a pretrained language model capable of interpreting SMILES and reaction-template descriptions, with Qwen3-4B offering the best trade-off between performance and computational cost.

\subsubsection{Computational Efficiency}
\label{sec:efficiency}

Environment-LLM inference is a major computational cost during RL exploration. The cache described in Section~\ref{sec:environment} stores proposed actions and reaction outcomes for each previously seen molecule. When GRPO revisits a previously seen molecule, these results are reused without another LLM call. Across all 14 tasks, 15{,}702 of 27{,}828 environment calls are cache hits, corresponding to a hit rate of 56.4\%. Including both LLM inference and policy training, end-to-end optimization takes 40 hours per task with caching, compared with an estimated 70 hours without caching. This reduces the overall computational cost by about 43\%, making multi-step policy optimization more efficient.

Beyond the efficiency gains from caching, we also compare the computational requirements across methods. All methods follow the same oracle-call budget, providing a basis for comparing sample efficiency, while computational costs are reported separately in Appendix~\ref{app:compute}. MolReAct instead fine-tunes a 4B policy model for each task, whereas synthesizable baselines typically require multi-day pretraining of dedicated models on large molecular datasets. MolReAct therefore avoids dedicated pretraining corpora while maintaining a per-task computational cost comparable to these baselines.

\begin{figure}[t]
\centering
\includegraphics[width=.85\linewidth]{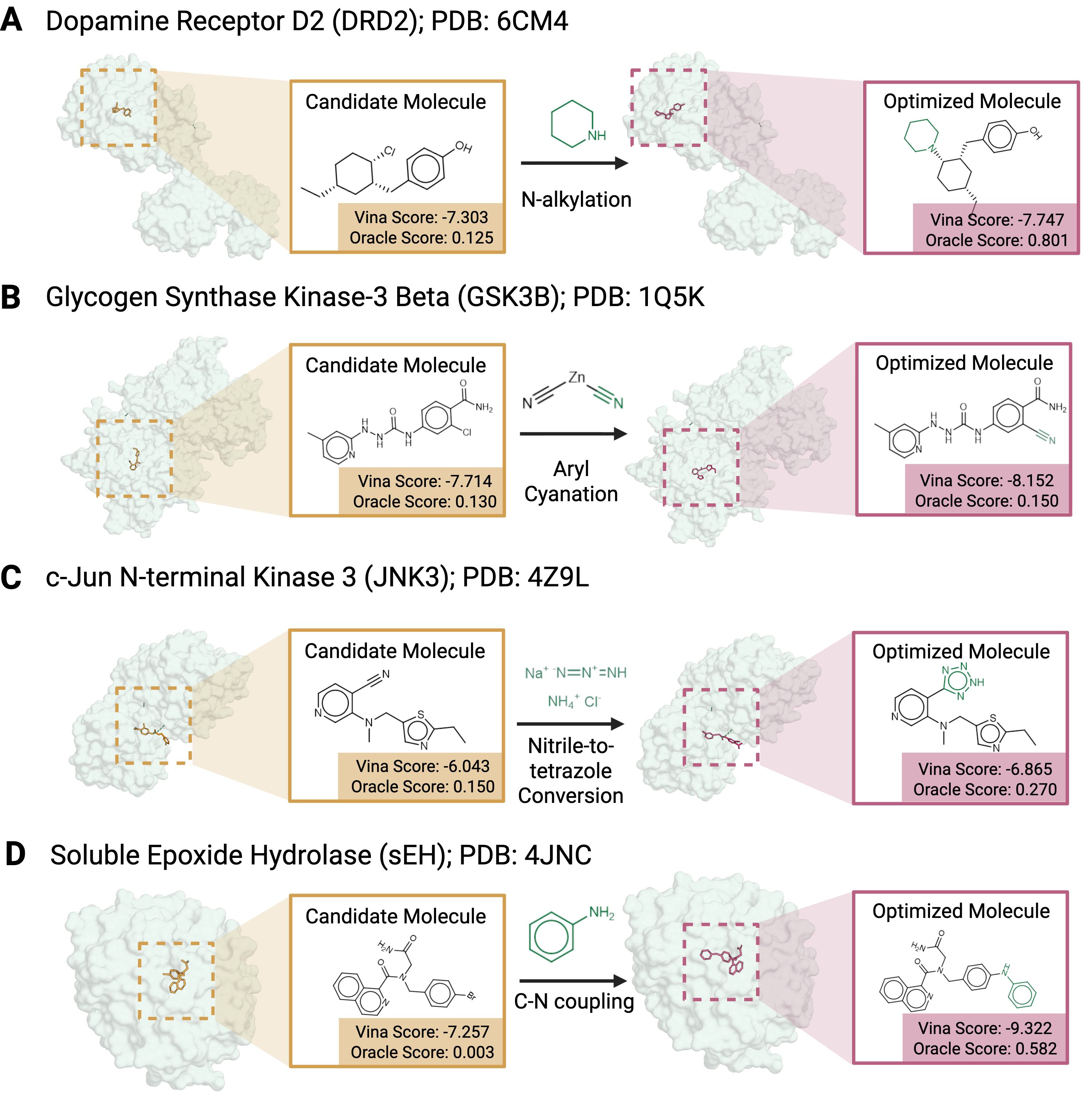}
\caption{Representative template-grounded reaction pathways discovered by MolReAct on four protein-target activity tasks. Each panel shows the candidate molecule, the applied template-grounded reaction with its building block, and the product with corresponding AutoDock Vina scores.}

\label{fig:case_studies}

\end{figure}

\subsubsection{Pathway Analysis}
\label{sec:chemical_validity}

Figure~\ref{fig:case_studies} presents four representative multi-step reaction pathways generated by MolReAct for DRD2, GSK3$\beta$, JNK3, and sEH. We additionally compute AutoDock Vina scores~\citep{autodock_vina} as a structure-based measure of target binding. Across the four targets, MolReAct identifies chemically meaningful transformations, including N-alkylation for DRD2, aryl cyanation for GSK3$\beta$, nitrile-to-tetrazole conversion for JNK3, and C--N coupling for sEH. These modifications consistently improve both the oracle and Vina scores. Beyond individual cases, the tool-augmented LLM proposes an average of 3.81 reactions during MolReAct's search (Figure~\ref{fig:action_dist}), while the optimized molecules are obtained through roughly two template-grounded reactions on average (Figure~\ref{fig:path_length}). This suggests that MolReAct constructs a compact action space while supporting effective multi-step decision-making.

\section{Discussion}

This paper presents MolReAct, a framework for synthesizable molecular optimization that uses a tool-augmented LLM to construct a compact action space and then integrates it with reinforcement learning for multi-step optimization. This differs from the protocol used in de novo synthesizable design, where methods such as SynthesisNet~\citep{PSSM} and GFlowNet-based generators~\citep{RGFN,SynFlowNet} assemble molecules from building blocks over a reaction graph without conditioning on an existing molecule. In contrast, MolReAct applies template-grounded edits directly to a given molecule, producing an explicit and traceable reaction pathway to each optimized product.

Nevertheless, several limitations remain. First, the quality of the action space at each step depends on the environment LLM, and weaker models may degrade proposal quality over multi-step trajectories. Beyond model capability, the purchasability filter ensures that selected building blocks are available from the Enamine catalog, but it may exclude useful reactions and remains limited by the commercial catalog coverage and exact-SMILES matching. On the transition side, template-grounded execution does not explicitly model reaction conditions or condition-dependent yields, and the resulting pathways therefore require experimental validation to confirm their real-world practical feasibility. Finally, each task optimizes a single scalar oracle, whereas practical drug-development workflows involve multiple competing objectives. Extending the terminal reward to multi-objective settings is therefore a natural direction for future work. More broadly, MolReAct is intended to support early-stage drug discovery. Its search is constrained to validated reaction templates and purchasable building blocks, prioritizing well-established transformations rather than less conventional chemistry.

\bibliography{references.bib}
\bibliographystyle{iclr2027_conference}

\newpage
\appendix
\newpage

\section{Implementation Details}
\label{app:implementation}

\subsection{GRPO Training Hyperparameters}

The policy model is Qwen3-4B-Instruct, trained with GRPO using the MemoryEfficientAdamW optimizer on a single NVIDIA RTX 6000 Ada GPU (48,GB), separate from the four GPUs serving the environment agent (Appendix~\ref{app:env_agent}).
Table~\ref{tab:hyperparams} summarizes the hyperparameters shared across all 14 benchmark tasks in our experiments.
A KL regularization term ($\beta_{\text{KL}}{=}0.02$) penalizes drift from a reference policy.
Each training batch consists of 30 seed molecules, each expanded into 10 trajectories, accumulated over a micro-batch of 20.
The policy is trained for 25 steps, sufficient for the small discrete classification at each step among at most 10 candidates.
The maximum reaction chain per trajectory is 5 steps, and the LLM agent is allowed up to 10 ReAct iterations per step. Training on a single property takes approximately 40 hours on the same GPU, including LLM inference.

\begin{table}[h]
\centering
\caption{GRPO hyperparameters for all benchmark tasks.}
\label{tab:hyperparams}
\small
\begin{tabular}{lc}
\toprule
\textbf{Hyperparameter} & \textbf{Value} \\
\midrule
Policy model                        & Qwen3-4B-Instruct \\
Training dtype                      & bfloat16 \\
Optimizer                           & MemoryEfficientAdamW \\
Adam $(\beta_1, \beta_2)$           & $(0.9,\ 0.999)$ \\
Weight decay                        & 0.0 \\
Learning rate                       & $1.0 \times 10^{-5}$ \\
Training steps                      & 25 \\
Clip ratio $\varepsilon$            & 0.2 \\
Discount factor $\gamma$            & 1.0 \\
KL coefficient $\beta_{\text{KL}}$  & 0.02 \\
KL reference policy                 & Fixed (initial) \\
Entropy coefficient                 & 0.01 \\
Molecules per batch                 & 30 \\
Trajectories per molecule           & 10 \\
Micro-batch size                    & 20 \\
Update epochs per rollout           & 2 \\
Max reaction depth                  & 5 \\
Agent max iterations                & 10 \\
\bottomrule
\end{tabular}
\end{table}

\subsection{Environment Agent Deployment}
\label{app:env_agent}

The environment agent runs Llama-3.3-70B-Instruct served via vLLM across four NVIDIA RTX 6000 Ada GPUs (48,GB each).
Inference uses a low decoding temperature $\tau{=}0.1$ to favor near-deterministic and stable reaction selection.
For each independent run, the reaction cache is initialized empty and populated anew, with no cache entries shared across runs.
For each step the agent receives the current SMILES, task-specific objective description and tool-call outputs, then proposes up to 10 candidate reactions.
Before template execution, we retain only proposals whose additional reactants exactly match canonical SMILES in the Enamine building-block catalog~\citep{enamine}.
The 115-template library and this candidate cap bound the reactions the agent can propose.

\subsection{Reaction Template Library}

The library contains 115 named organic reactions, each encoded as a SMARTS string specifying the reactant pattern and product transformation.
The template set is identical to that used by SynFormer~\citep{Synformer} and ReaSyn~\citep{Reasyn}, ensuring direct comparability with the synthesis-aware baselines.
At each step, applicable templates are identified by matching the molecule against every reactant SMARTS via RDKit \texttt{HasSubstructMatch}~\citep{rdkit}, and only templates that pass this filter are exposed to the agent.
The library spans major reaction classes: C-C bond formation, heteroatom introduction, ring closure, and functional-group interconversion.
When a template execution yields multiple products (\emph{e.g.}, from repeated reactant matches), we canonicalize each output via RDKit \texttt{MolToSmiles} and retain the first valid product in the enumeration order returned by \texttt{RunReactants}. This deterministic tie-break ensures that the same (state, template, building block) tuple always produces the same successor molecule, enabling the SMILES-based caching in Section~\ref{sec:environment}.
The full SMARTS list and reaction-execution scripts are included in our source code, which we provide to reviewers during the discussion period.

\begin{wraptable}{r}{0.53\textwidth}
\vspace{-1\baselineskip}
\centering
\caption{Ablation results on four protein-target activity tasks. Panel (a) evaluates single-step reaction proposals, and panel (b) evaluates multi-step optimization policies. The learned-policy results are evaluated using final checkpoint, so their values are not directly comparable to those in Table~\ref{tab:top10} and Table~\ref{tab:policy_scale}.}
\label{tab:ablation_numbers}

\small
\setlength{\tabcolsep}{4pt}
\renewcommand{\arraystretch}{1.1}

\begin{subtable}[t]{\linewidth}
\centering
\caption{Single-step reaction proposal variants.}
\begin{tabular}{lcccc}
\toprule
& DRD2 & GSK3$\beta$ & JNK3 & sEH \\
\midrule
Random Template & 0.233 & 0.294 & 0.194 & 0.615 \\
Vanilla LLM & 0.561 & 0.311 & 0.212 & 0.661 \\
Tool-LLM & \textbf{0.764} & \textbf{0.345} & \textbf{0.316} & \textbf{0.725} \\
\bottomrule
\end{tabular}
\end{subtable}

\vspace{0.2cm}

\begin{subtable}[t]{\linewidth}
\centering
\caption{Multi-step optimization policies.}
\begin{tabular}{lcccc}
\toprule
& DRD2 & GSK3$\beta$ & JNK3 & sEH \\
\midrule
Random Policy & 0.613 & 0.305 & 0.262 & 0.632 \\
Greedy Search & 0.819 & 0.622 & 0.413 & 0.682 \\
MCTS & 0.814 & 0.651 & 0.461 & 0.716 \\
MolReAct (GRPO) & \textbf{0.972} & \textbf{0.710} & \textbf{0.527} & \textbf{0.874} \\
\bottomrule
\end{tabular}
\end{subtable}

\vspace{-1\baselineskip}
\end{wraptable}

\begin{figure}[h]
\centering
\begin{subfigure}[b]{0.49\textwidth}
  \centering
  \includegraphics[width=\linewidth]{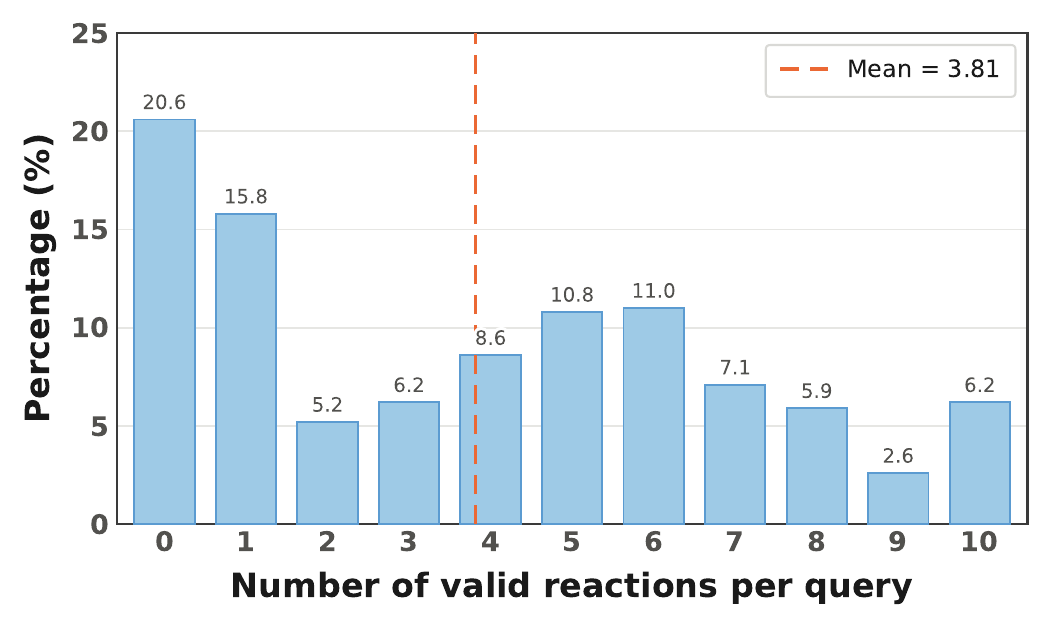}
  \caption{Distribution of valid reactions proposed per query.}
  \label{fig:action_dist}
\end{subfigure}
\hfill
\begin{subfigure}[b]{0.49\textwidth}
  \centering
  \includegraphics[width=\linewidth]{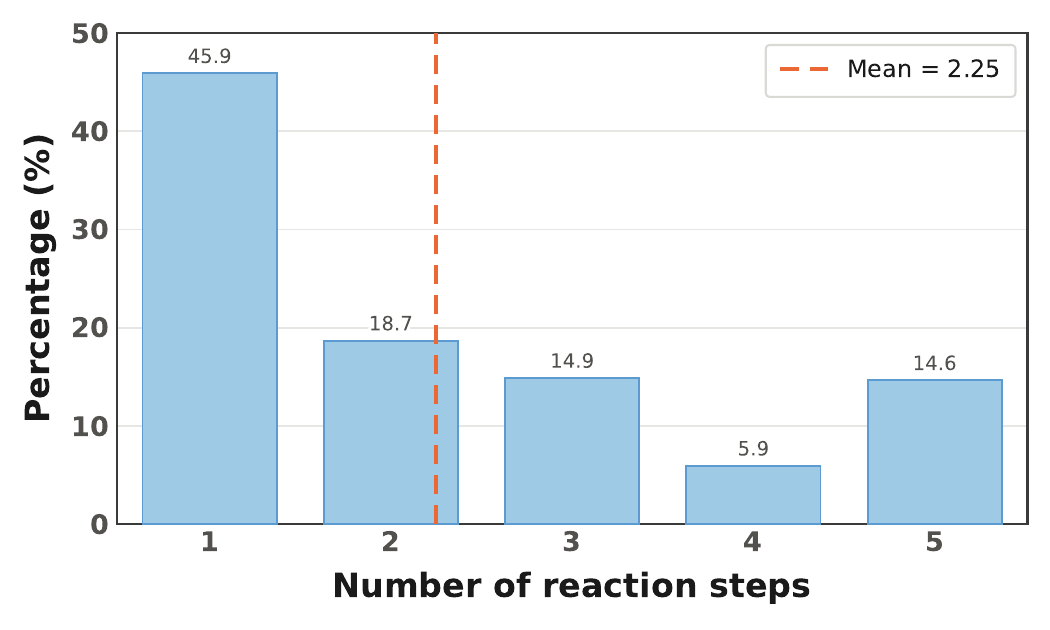}
  \caption{Distribution of reaction-path lengths.}
  \label{fig:path_length}
\end{subfigure}
\caption{Distributions characterizing MolReAct's search on the four protein-target activity tasks. (a) Valid reactions proposed per query during the search process. (b) Reaction-path lengths, the number of reactions applied before a trajectory ends at the stop action or the depth-5 limit.}
\label{fig:search_structure}
\end{figure}

\subsection{Task-Specific Information}
\label{app:task_information}

We ask whether the performance of MolReAct depends on being told what each task optimizes. The anonymized variants leave the \texttt{property\_description} field empty in the Question Prompt of Appendix~\ref{app:prompt}, removing both the task name and the natural-language objective description, so the environment agent sees only the current SMILES, the matched reaction templates, and the tool outputs. Everything else is identical to the corresponding full configuration: the same Llama-3.3-70B environment model and decoding temperature, the Enamine building-block filter, the maximum depth of 5, and the oracle budget of Appendix~\ref{app:oracle_budget}. For the anonymized multi-step framework the description is withheld during both policy training and evaluation, so the policy is never exposed to it, while the terminal oracle reward is unchanged. Because these policies are evaluated at their final checkpoint, their values are not directly comparable to Table~\ref{tab:top10}.

Table~\ref{tab:task_information} reports both settings on all 14 benchmark tasks. Averaged over the tasks, the anonymized agent retains $94.8\%$ of the one-step proposal quality and $98.8\%$ of the multi-step optimization performance. The objective description therefore contributes a small but consistent gain to single-step proposals, and almost none to the full framework, indicating that the learned policy recovers most of the information the description provides rather than by explicit task information.

\begin{table}[h]
\centering
\caption{Top-10 quality with and without task-specific objective descriptions across all 14 tasks. Panel (a) evaluates single-step reaction proposals, and panel (b) evaluates the full multi-step framework, whose policies are evaluated using the final checkpoint. $\Delta$ denotes Task-Specific minus Anonymous.}
\label{tab:task_information}
\small
\setlength{\tabcolsep}{4pt}
\renewcommand{\arraystretch}{1.08}
\begin{tabular}{lccccccc}
\toprule
& \multicolumn{3}{c}{\textbf{(a) One-step proposal}} & & \multicolumn{3}{c}{\textbf{(b) Multi-step optimization}} \\
\cmidrule(lr){2-4} \cmidrule(lr){6-8}
\textbf{Task} & \textbf{Anon.} & \textbf{Task-Spec.} & \textbf{$\Delta$} & & \textbf{Anon.} & \textbf{Task-Spec.} & \textbf{$\Delta$} \\
\midrule
\multicolumn{8}{l}{\textit{Multi-Parameter Optimization}} \\
\midrule
amlodipine\_mpo   & 0.430 & \textbf{0.462} & +0.032 & & 0.442 & \textbf{0.474} & +0.032 \\
fexofenadine\_mpo & 0.699 & \textbf{0.742} & +0.043 & & \textbf{0.759} & 0.756 & -0.003 \\
osimertinib\_mpo  & 0.727 & \textbf{0.775} & +0.048 & & 0.802 & \textbf{0.812} & +0.010 \\
perindopril\_mpo  & 0.408 & \textbf{0.438} & +0.030 & & 0.447 & \textbf{0.456} & +0.009 \\
ranolazine\_mpo   & 0.697 & \textbf{0.749} & +0.052 & & 0.781 & \textbf{0.794} & +0.013 \\
sitagliptin\_mpo  & 0.296 & \textbf{0.337} & +0.041 & & 0.362 & \textbf{0.371} & +0.009 \\
zaleplon\_mpo     & \textbf{0.427} & 0.423 & -0.004 & & 0.436 & \textbf{0.452} & +0.016 \\
\midrule
\multicolumn{8}{l}{\textit{Rediscovery and Median}} \\
\midrule
celecoxib\_rediscovery & \textbf{0.351} & 0.348 & -0.003 & & 0.375 & \textbf{0.382} & +0.007 \\
median\_1              & 0.181 & \textbf{0.218} & +0.037 & & \textbf{0.264} & 0.243 & -0.021 \\
median\_2              & 0.130 & \textbf{0.164} & +0.034 & & \textbf{0.171} & 0.167 & -0.004 \\
\midrule
\multicolumn{8}{l}{\textit{Protein Binding Activity}} \\
\midrule
DRD2        & 0.752 & \textbf{0.764} & +0.012 & & 0.961 & \textbf{0.972} & +0.011 \\
GSK3$\beta$ & 0.331 & \textbf{0.345} & +0.014 & & \textbf{0.712} & 0.710 & -0.002 \\
JNK3        & 0.313 & \textbf{0.316} & +0.003 & & 0.513 & \textbf{0.527} & +0.014 \\
sEH         & 0.707 & \textbf{0.725} & +0.018 & & 0.873 & \textbf{0.874} & +0.001 \\
\midrule
Average & 0.461 & \textbf{0.486} & +0.025 & & 0.564 & \textbf{0.571} & +0.007 \\
\bottomrule
\end{tabular}
\end{table}

\subsection{Oracle Call Accounting}
\label{app:oracle_budget}

The 10{,}000-call budget of \citet{gao_benchmark} accounts for every oracle query made during policy training and evaluation. At each of the 25 training steps, we sample 30 molecules and generate 10 trajectories per molecule, resulting in 300 terminal oracle calls. We then evaluate the policy by generating one terminal product from each of the same 100 held-out starting molecules used by all methods, adding 100 evaluation calls. This gives 7{,}500 training calls and 2{,}500 evaluation calls within the budget of 10{,}000. Products generated during training are excluded from all reported metrics. For each run, the Top-10 score is computed over the 2{,}500 terminal products accumulated from the held-out evaluations. AUC$_{10}$ is computed from the cumulative Top-10 score over all held-out evaluation products generated up to each training step. Its horizontal axis counts both training and evaluation oracle calls, reaching 10{,}000 calls after 25 steps. The comparison therefore charges MolReAct for the cost of learning its policy rather than treating training as free.

\newpage

\subsection{Fingerprint Policy Baseline}
\label{app:ecfp_policy}

Since selecting among at most 10 candidates is a small discrete choice, we ask whether the language-model policy can be replaced by a lightweight fingerprint-based alternative. Each candidate reaction at state $s_t$ is described by the ECFP4 fingerprint of the product (2048 bits, Morgan radius 2), the difference between the product and state fingerprints, and a one-hot encoding of the reaction template over the 115-template library, giving a 4211-dimensional feature vector. A two-layer MLP with hidden widths 512 and 256 maps each feature vector to one logit, for 2.29M parameters against 4B for Qwen3-4B, with weights shared across candidates so that the variable-size action space is handled directly. Unused slots are masked with $-\infty$ and a softmax gives $\pi_\theta(a_t \mid s_t)$ as in Section~\ref{sec:rl_grpo}, so this policy is trained against the same terminal reward. The environment agent remains the frozen Tool-LLM configuration, and the Enamine catalog filter, template library, cap of 10 candidates per step, maximum depth of 5, and oracle budget of Appendix~\ref{app:oracle_budget} are all unchanged.

\section{Computing Resource Comparison}
\label{app:compute}

The costs of MolReAct and of the baselines are not incurred in the same way in practice, so we separate them rather than report a single figure. The baselines pay a large one-time cost to pretrain a model that is then reused across all 14 tasks, whereas MolReAct pretrains nothing but trains a policy separately for each task, so its cost scales with the number of tasks. Table~\ref{tab:compute} therefore reports each method's cost as its own paper does, and the comparison below places them on a common basis by amortizing each one-time cost over the 14 benchmark tasks. Figures are taken from each baseline's original publication; entries marked ``--'' were not reported and we did not estimate them.

\paragraph{Observations.}
Amortized over the 14 tasks, ReaSyn's pretraining comes to roughly 690 GPU-h per task and SynFormer's to roughly 82, against 200 GPU-h per task for MolReAct. MolReAct is thus about an order of magnitude cheaper than ReaSyn and roughly twice as expensive as SynFormer, and because nothing is amortized, its total cost grows linearly with the number of tasks, reaching about 2{,}800 GPU-h across the full benchmark, with four of the five GPUs occupied by the frozen 70B environment model. We therefore do not claim a compute advantage over the synthesis-constrained baselines. What the table does show is a different cost profile. MolReAct trains no model from scratch and needs no reaction-pathway dataset, so a new objective requires only a 4B policy fine-tune on commodity 48\,GB GPUs, with no 8$\times$A100 pretraining run and no 120-GPU data-generation stage. Its recurring cost is also reducible in ways the pretraining cost is not: the environment LLM is frozen and used only for inference, so it can be served from a shared or hosted endpoint across concurrent tasks rather than held on dedicated hardware. Because these profiles are not commensurable, the oracle-call budget of \citet{gao_benchmark}, which every method is held to identically, remains the fairness anchor for the comparisons in Section~\ref{sec:exp_settings}.

\begin{table}[h]
\centering
\caption{Computing resources across baselines and MolReAct. Reported training cost is one-time pretraining for the baselines, which is reused across all 14 tasks, and per-task training for MolReAct, which is not. Search cost is excluded for all methods, as it is not reported consistently.}
\label{tab:compute}
\setlength{\tabcolsep}{4pt}
\renewcommand{\arraystretch}{1.5}
\resizebox{\textwidth}{!}{%
\begin{tabular}{lll}
\toprule
\textbf{Method} & \textbf{Training framework (size)} & \textbf{Reported training cost} \\
\midrule
Graph GA & Genetic algorithm (no neural model) & none \\
ReaSyn & AR transformer (166M) + Edit Bridge (174M) & 5d on 8$\times$A100 + 3d data gen on 120$\times$A100 \\
SynFormer & Encoder-decoder transformer (229M) & 6d on 8$\times$A100 \\
DrugAssist & Llama2-7B-Chat with LoRA (7B) & -- (10 epochs, batch 512) on 8$\times$A100-40GB \\
LDMol & BERT-base AE + DiT-base ($\sim$250M) & -- on 8$\times$A100-40GB \\
mCLM & Qwen2.5-3B + GNN (3B) & -- (6 epochs pretrain, 5 epochs finetune) on 4$\times$A100-80GB \\
\midrule
MolReAct (Ours) & Qwen3-4B policy (4B) & 40 h on 5$\times$RTX 6000 Ada (per task, no pretraining) \\
\bottomrule
\end{tabular}%
}
\end{table}

\section{Monte Carlo Tree Search Details}
\label{app:mcts}

We implement Monte Carlo Tree Search (MCTS) as a planning baseline over the same Tool-LLM-generated reaction environment proposal provided by MolReAct. Each tree node represents a molecule, and its outgoing edges correspond to the proposed template-grounded reactions together with the stop action. During each simulation, tree selection follows the Upper Confidence Bound for Trees (UCT) criterion, $Q(s,a)+c\sqrt{\log N(s)/(N(s,a)+1)}$, where $Q(s,a)$ is the mean terminal reward backed up through action $a$, $N(s)$ and $N(s,a)$ are the corresponding visit counts, and the exploration constant is fixed to $c=\sqrt{2}$. After expanding an unvisited action, MCTS performs a uniform random rollout until the stop action is selected, the maximum reaction depth of five is reached, or no valid reaction is available. Only the terminal molecule is evaluated by the task oracle, and its reward is propagated through all nodes visited during the simulation.

We use the same MCTS configuration for DRD2, GSK3$\beta$, JNK3, and sEH, without task-specific hyperparameter tuning. For each task, MCTS runs 100 simulations independently from each of the 100 held-out starting molecules, resulting in 10,000 terminal oracle calls. Each state contains at most 10 reaction actions in addition to the stop action, matching the action-space configuration used by MolReAct. After search, we retain the highest-scoring terminal molecule discovered from each starting molecule and compute the Top-10 score over the resulting 100 outputs. Results are averaged over three independent MCTS seeds. MCTS uses neither a learned policy prior nor a learned value function; all search decisions are determined by UCT statistics and terminal oracle rewards.

\begin{table}[h]
\centering
\caption{Paired Wilcoxon signed-rank test of MolReAct against each baseline across 14 tasks. Wins/losses count tasks where the MolReAct score exceeds/trails the baseline.}
\label{tab:wilcoxon}
\small
\setlength{\tabcolsep}{4pt}
\renewcommand{\arraystretch}{1.05}
\begin{tabular}{lcccccc}
\toprule
& \multicolumn{3}{c}{\textbf{Top-10}} & \multicolumn{3}{c}{\textbf{AUC$_{10}$}} \\
\cmidrule(lr){2-4} \cmidrule(lr){5-7}
\textbf{vs Baseline} & W/L & two-sided $p$ & one-sided $p$ & W/L & two-sided $p$ & one-sided $p$ \\
\midrule
Graph GA   & 13/1  & $0.003$  & $0.002$  & 14/0 & $<0.001$ & $<0.001$ \\
ReaSyn     & 12/2  & $<0.001$ & $<0.001$ & 12/2 & $<0.001$ & $<0.001$ \\
SynFormer  & 13/1  & $<0.001$ & $<0.001$ & 13/1 & $0.001$  & $<0.001$ \\
DrugAssist & 14/0  & $<0.001$ & $<0.001$ & --   & --       & --       \\
LDMol      & 14/0  & $<0.001$ & $<0.001$ & --   & --       & --       \\
mCLM       & 14/0  & $<0.001$ & $<0.001$ & --   & --       & --       \\
\bottomrule
\end{tabular}
\end{table}

\section{Statistical Significance}
\label{app:significance}

To assess whether the per-task improvements reflect systematic gains rather than run-to-run variance, we conduct paired Wilcoxon signed-rank tests across all 14 benchmark tasks. Top-10 comparisons include all six baselines in Table~\ref{tab:top10}, whereas AUC$_{10}$ comparisons include the three baselines reported in Table~\ref{tab:auc10}. Table~\ref{tab:wilcoxon} reports both two-sided and one-sided $p$-values (testing the directional hypothesis that MolReAct exceeds the baseline), together with per-task win/loss counts.

\paragraph{Observations.}
MolReAct significantly outperforms all six baselines on Top-10 and all three reported baselines on AUC$_{10}$ (two-sided $p<0.05$). The gain over the strongest synthesis-constrained baseline, SynFormer, is significant on both metrics (13/1 wins, two-sided $p<0.001$). 


\begin{table}[h]
\centering
\caption{Optimization objectives for all 14 benchmark tasks.}
\label{tab:property_descriptions}
\small
\resizebox{\textwidth}{!}{%
\begin{tabular}{@{}lp{12cm}@{}}
\toprule
\textbf{Property} & \textbf{Objective} \\
\midrule
\multicolumn{2}{l}{\textit{Multi-Parameter Optimization (MPO)}} \\
\midrule
Amlodipine MPO    & Maximize similarity to Amlodipine; keep total ring count $\approx$3. \\
Fexofenadine MPO  & Maximize similarity to Fexofenadine; raise TPSA ($\geq90$); keep logP $\leq4$. \\
Osimertinib MPO   & Maximize similarity to Osimertinib; raise TPSA ($\geq100$); push logP very low ($\leq1$). \\
Perindopril MPO   & Maximize similarity to Perindopril; match aromatic ring count $\approx2$. \\
Ranolazine MPO    & Maximize similarity to Ranolazine; raise TPSA ($\geq95$) and logP ($\geq7$); keep fluorine count $\approx1$. \\
Sitagliptin MPO   & Maximize dissimilarity to Sitagliptin; match logP $\approx2$ and TPSA $\approx77$. \\
Zaleplon MPO      & Maximize fingerprint similarity to Zaleplon. \\
\midrule
\multicolumn{2}{l}{\textit{Rediscovery \& Median}} \\
\midrule
Celecoxib Rediscovery & Recover the exact scaffold and substituent pattern of Celecoxib via fingerprint similarity. \\
Median 1          & Maximize simultaneous similarity to camphor and menthol (terpene-like intermediate). \\
Median 2          & Maximize simultaneous similarity to tadalafil and sildenafil (multi-ring heteroaromatic intermediate). \\
\midrule
\multicolumn{2}{l}{\textit{Protein Binding Activity}} \\
\midrule
JNK3              & Maximize predicted inhibitory activity against JNK3 kinase. \\
DRD2              & Maximize predicted activity against DRD2 (aminergic GPCR). \\
GSK3$\beta$       & Maximize predicted inhibitory activity against GSK3$\beta$ kinase. \\
sEH               & Maximize predicted binding affinity for soluble epoxide hydrolase (sEH/EPHX2). \\
\bottomrule
\end{tabular}%
}
\end{table}

\newpage

\section{Benchmark Property Descriptions}
\label{app:properties}

The benchmark includes 13 property optimization tasks from TDC and one proxy-based binding affinity task for soluble epoxide hydrolase (sEH). Table~\ref{tab:property_descriptions} summarizes their optimization objectives across MPO, rediscovery and median objectives, and protein-target activity prediction.

\section{Chemical Tool Details}
\label{app:tools}

All six tools are implemented with RDKit (v2024.03) and registered as LangChain \texttt{BaseTool} instances~\citep{langchain}.
Each tool takes the current molecule SMILES as input and runs locally, returning a structured plain-text string for the agent to reason over in the ReAct scratchpad.
Table~\ref{tab:tools_extended} lists each tool alongside its chemical role and a representative output.
\texttt{SiteAnalyzer} detects five reactive motifs: electrophilic carbonyl, nucleophilic amine, alcohol/phenol, halogen, and Michael acceptor.
Together, these tools provide the agent with a compact structural profile sufficient to select structurally compatible reaction templates without querying any external service.

\begin{table}[h]
\centering
\caption{Chemical analysis tools for LLM agent. All accept the SMILES and return structured text.}
\label{tab:tools_extended}
\small
\resizebox{\textwidth}{!}{%
\begin{tabular}{@{}lll@{}}
\toprule
\textbf{Tool} & \textbf{Role} & \textbf{Example output} \\
\midrule
\texttt{SMILES2Weight}   & Molecular weight             & \texttt{335.14} \\[4pt]
\texttt{FuncGroups}       & Functional group detection   & \texttt{carboxylic acid (1), primary amine (1)} \\[4pt]
\texttt{ScaffoldExtract}  & Murcko scaffold extraction   & \texttt{[Scaffold] c1ccc2ncccc2c1; rings=2; hetero=1} \\[4pt]
\texttt{BRICSFragment}    & BRICS decomposition          & \texttt{frag\_count=3; frags=[CCN, c1ccncc1, C(=O)O]} \\[4pt]
\texttt{RingAnalyzer}     & Ring analysis                & \texttt{total=3; arom=2; hetero=1; [6A,6A,5H]} \\[4pt]
\texttt{SiteAnalyzer}     & Reactive site identification & \texttt{count=2; sites=[carbonyl, nucl\_amine]} \\
\bottomrule
\end{tabular}%
}
\end{table}

\clearpage
\raggedbottom
\section{LLM Agent Prompt}
\label{app:prompt}

\begin{tcolorbox}[
    enhanced,
    colback=white,
    colframe=black!60,
    boxrule=0.8pt,
    arc=4pt,
    left=1pt, right=1pt, top=2pt, bottom=2pt, 
    width=\linewidth,
    breakable,
    before skip=2pt,
]

\begin{tcolorbox}[
    colback=blue!3, colframe=blue!25,
    title={\footnotesize\textbf{System Role}},
    fonttitle=\footnotesize\bfseries, coltitle=blue!50!black,
    boxrule=0.5pt, arc=3pt, left=2pt, right=2pt, top=2pt, bottom=2pt,
]
\scriptsize\ttfamily
You are an expert medicinal chemist.
\end{tcolorbox}

\vspace{3pt}

\begin{tcolorbox}[
    colback=green!5, colframe=green!45!black,
    title={\footnotesize\textbf{Format Instructions}},
    fonttitle=\footnotesize\bfseries, coltitle=white,
    boxrule=0.5pt, arc=3pt, left=2pt, right=2pt, top=2pt, bottom=2pt,
    breakable,
]
\scriptsize\ttfamily
You can respond in two ways:\\[3pt]
\textbf{Thought:} (reflect on your progress and decide what to do next)\\
\textbf{Action:} (the action name, should be one of [\textcolor{purple}{\{tool\_names\}}])\\
\textbf{Action Input:} (the input string to the action)\\[3pt]
OR\\[3pt]
\textbf{Final Answer:}\\[3pt]
**Reaction 1:**\\
- Reaction name: [reaction name[id] from the feasible template list above]\\
- Reactants: [ONLY the missing reactant(s) SMILES --- see rules below]\\[3pt]
**Reaction 2:**\\
- Reaction name: [reaction name[id]]\\
- Reactants: [ONLY the missing reactant(s) SMILES]\\[3pt]
{[}List up to 10 reactions from the templates provided.
Choose based on the provided optimization objective. Use a DIVERSE set of
templates. If reusing the same template, you MUST use a different reactant each time.{]}\\[4pt]
\begin{tcolorbox}[
    colback=orange!8, colframe=orange!50,
    title={\footnotesize\textbf{Critical Rules}},
    fonttitle=\footnotesize\bfseries, coltitle=orange!60!black,
    boxrule=0.4pt, arc=2pt, left=2pt, right=2pt, top=2pt, bottom=2pt,
]
\scriptsize\ttfamily
(1) Reaction name must be EXACTLY one of the templates listed above.\\[2pt]
(2) Reactants field --- STRICT requirements:\\
\quad- ONLY the missing/additional reactant SMILES. NEVER include the input molecule.\\
\quad- The number of reactants MUST match the template's missing count.\\
\quad- Each reactant must be COMPLETE, VALID with $\geq$2 heavy atoms.
  Single atoms (O, N, C, Cl, Br) are \textbf{FORBIDDEN}.\\
\quad- If the template says ``reactants: none'', write ``Reactants: none''.\\[2pt]
(3) How to read SMARTS and pick the right molecule:\\
\quad (a) identify the atom chain; (b) build a real molecule; (c) close any aromatic rings.\\[2pt]
\textit{Common SMARTS $\rightarrow$ reactant examples:}\\
\quad\texttt{[O\&H1\$(Oc)]} (phenol) $\rightarrow$ \texttt{Oc1ccccc1} \quad
\texttt{[N\&X3;H1,H2\$(N[\#6])]} (amine) $\rightarrow$ \texttt{NCC, NC(C)C}\\
\quad\texttt{[\#6][C\&H1]=O} (aldehyde) $\rightarrow$ \texttt{CC=O} \quad
\texttt{[\#6]C(=O)[O\&H1]} (acid) $\rightarrow$ \texttt{CC(=O)O} \quad
\texttt{[Cl,Br,I][\#6]} $\rightarrow$ \texttt{ClCC}\\[2pt]
Never output a bare atom. Always output a full molecule.
\end{tcolorbox}
\vspace{3pt}
\scriptsize\ttfamily
\textbf{Important:} You may use tools to understand the molecule. After every Observation,
write a Thought, then either Action + Action Input, or Final Answer.
Use exactly ``\textbf{Final Answer:}'' before **Reaction 1**.
\end{tcolorbox}

\vspace{3pt}

\begin{tcolorbox}[
    colback=violet!5, colframe=violet!40,
    title={\footnotesize\textbf{Question Prompt}},
    fonttitle=\footnotesize\bfseries, coltitle=violet!60!black,
    boxrule=0.5pt, arc=3pt, left=2pt, right=2pt, top=2pt, bottom=2pt,
]
\scriptsize\ttfamily
Answer the question below using the following Tools:\\[3pt]
\textcolor{purple}{\{tool\_strings\}}\\[3pt]
Question:\\
Input molecule (SMILES): \textcolor{purple}{\{smiles\}}\\
Optimization objective: \textcolor{purple}{\{property\_description\}}\\
Design up to 10 diverse, chemically feasible single-step reactions.\\
Reaction templates: \textcolor{purple}{\{matched\_templates\}}
\end{tcolorbox}

\end{tcolorbox}


\end{document}